\pdfoutput=1

\documentclass[11pt]{article}

\usepackage[preprint]{acl}
\usepackage{placeins}
\usepackage{float}
\usepackage{amsthm}
\usepackage{times}
\usepackage{latexsym}

\usepackage[T1]{fontenc}
\usepackage{enumitem}
\usepackage{amsmath}
\usepackage[utf8]{inputenc}

\usepackage{microtype}

\usepackage{inconsolata}

\usepackage{graphicx}
\usepackage[nolist]{acronym}

\usepackage{amssymb}

\usepackage{booktabs}
\usepackage{caption}     
\usepackage{makecell} 
\usepackage{graphicx} 
\usepackage{footnote} 
\usepackage{tabularx}
\usepackage{physics}
\begin{acronym}
\acro{llm}[LLM]{Large Language Model}
\acro{llms}[LLMs]{Large Language Models}
\acro{lms}[LMs]{Language Models}
\acro{lm}[LM]{Language Model}
\acro{mi}[MI]{Mechanistic Interpretability}
\acro{sae}[SAE]{Sparse Autoencoder}
\acro{saes}[SAEs]{Sparse Autoencoders}
\end{acronym}
\title{Interpretable Company Similarity with Sparse Autoencoders\thanks{This work appeared as a preprint on arXiv:  \\  \url{https://arxiv.org/abs/2412.02605}. \\  Code and data are available at: \url{https://github.com/FlexCode29/company_similarity_sae}. \\ Alternative email: \href{mailto:marcomolinari4@gmail.com}{marcomolinari4@gmail.com}

}}

\author{
  Marco Molinari\textsuperscript{\textdagger, 1} \and Victor Shao\textsuperscript{\textdagger, 1} \and \textbf{Luca Imeneo}\textsuperscript{2} \\ \and \textbf{Mateusz Mikolajczak}\textsuperscript{1} \and \textbf{Vladimir Tregubiak}\textsuperscript{1} \\ 
  \and \textbf{Abhimanyu Pandey}\textsuperscript{1} \and \textbf{Sebastião Kuznetsov Ryder Torres Pereira}\textsuperscript{1} \\[1em]
  \textsuperscript{1} LSE.AI, London School of Economics\\
  \textsuperscript{2} Tower Research Capital\\[1em]
  \textbf{Correspondence:} \href{mailto:m.molinari1@lse.ac.uk}{m.molinari1@lse.ac.uk}\hyperlink{firstpage}{\textsuperscript{*}} \\
  \textsuperscript{\textdagger} Equal contribution
}

\begin{document}
\maketitle
\begin{abstract}
Determining company similarity is a vital task in finance, underpinning risk management, hedging, and portfolio diversification. Practitioners often rely on sector and industry classifications such as SIC and GICS codes to gauge similarity, the former is used by the U.S. Securities and Exchange Commission (SEC), and the latter widely used by the investment community. Since these classifications lack granularity and need regular updating, using clusters of embeddings of company descriptions has been proposed as a potential alternative, but the lack of interpretability in token embeddings poses a significant barrier to adoption in high-stakes contexts. \ac{saes} have shown promise in enhancing the interpretability of \ac{llms} by decomposing \ac{llm} activations into interpretable features. Moreover, \ac{saes} capture an \ac{llm}'s internal representation of a company description, as opposed to semantic similarity alone, as is the case with embeddings. We apply \ac{saes} to company descriptions, and obtain meaningful clusters of equities. We benchmark SAE features against SIC-codes, Industry codes, and Embeddings. Our results demonstrate that SAE features surpass sector classifications and embeddings in capturing fundamental company characteristics. This is evidenced by their superior performance in correlating logged monthly returns – a proxy for similarity – and generating higher Sharpe ratios in co-integration trading strategies, which underscores deeper fundamental similarities among companies. Finally, we verify the interpretability of our clusters, and demonstrate that sparse features form simple and interpretable explanations for our clusters.
\end{abstract}

\hypertarget{firstpage}{}

\section{Introduction}

Accurately assessing the similarity of companies is an integral task in finance, key to risk management, portfolio diversification and more \citep{Delphini[1], Katselas[1b]}. Hedging, a practice that relies on converse investments in related assets, is a prominent example of a financial strategy that requires a detailed understanding of the similarity between two companies.

Traditionally, company comparisons rely on (1) relative returns and (2) discrete classifications, or a combination of both\footnote{E.g. SIC-codes \citep{SICmanual645}, and the Global Industry Classification System (GICS), which categorizes companies into 11 sectors and 163 sub-industries \citep{MSCI:GICS2020[3]}. }. For the former, relying on relative return spreads can be effective but is not foolproof, as market volatility, economic changes, fundamental changes in business, and temporal factors can alter them \citep{Loretan[4]}. For the latter, discrete classification systems such as GICS\footnotemark[1] are limited, as the restricted granularity of a discrete classification system limits dynamic interpretations of companies' operations, in that they fail to account for the duality of certain companies \footnote{Emerging industries disproportionally exhibit this.} \citep{Winton[5]}. 

This is particularly important for pairs trading, a market-neutral strategy based on mean-reverting return spreads \citep{Pairs[7]}. Employing a pair-trading strategy with fundamentally similar companies whose returns are co-integrated\footnote{Co-integration refers to a statistical property where two or more non-stationary time series variables, despite individual trends, exhibit a stationary linear combination, indicating a  long-term equilibrium relationship \citep{EngleGranger}.} could reduce the risk of deviation from historical trends \citep{bharadwaj2014pairs}.

Clustering embeddings of company descriptions has been proposed as a measure of similarity \citep{vamvourellis2023company, buchner-etal-2024-prompt}, but token embeddings are not interpretable, and this leads to uncertainty, which is undesirable in the financial sector.

\ac{saes} have the potential to provide an efficient measure of company similarity by decomposing large amounts of financial data into interpretable features \citep{9378325}. \ac{saes} have recently been applied to \ac{llms} resulting in interpretable decompositions of neural activations \citep{cunningham2023sparseautoencodershighlyinterpretable}. Furthermore, \ac{saes} can be applied at a \ac{lm}'s deeper layers, and hence decompose a \ac{lm}'s internal representation of a company description, which means \ac{sae} features capture more abstract and cross-token concepts than raw embeddings \citep{templeton2024scaling}. This motivates their application to textual company descriptions.

To the best of our knowledge, we are the first to compute company similarity using \ac{saes} on SEC\footnote{Securities and Exchange Commission} filings, and to show that \ac{saes} can surpass existing alternatives on identifying similar companies despite the sparsity (interpretability) constraint. This is relevant since the competitiveness of \ac{saes} has been called into question \citep{kantamneni2025sparseautoencodersusefulcase} when compared with existing benchmarks of downstream performances.

Our contributions can be summarized as follows:
\begin{itemize}
    \item We apply an open source SAE \citep{eleutherai_sae_llama} to Llama 3.1 8B \citep{dubey2024llama3herdmodels}, and release a dataset containing company descriptions, extracted features, and returns, to support further research.\hyperlink{firstpage}{\textsuperscript{*}}
    \item We demonstrate that clustering using sparse features outperforms embeddings and SIC/GISC codes \citep{MSCI:GICS2020[3]} in terms of intra-cluster pairwise correlations.
    \item We confirm the interpretability of our clusters by verifying that our explanations use a small number of highly interpretable features.
\end{itemize}
\section{Related Works}
\subsection{Sparse autoencoders}

The Linear Representation Hypothesis posits that \ac{llms} linearly represent concepts in neuron activations \citep{park2024linearrepresentationhypothesisgeometry}. However, as neuron activations are notoriously superpositioned \citep{elhage2022toymodelssuperposition}, \ac{saes} enhance the interpretability of \ac{llms} by writing neuron activations as a linear combination of sparse features \citep{SAEs_fix_it}. This reduces superposition and restores interpretability \citep{cunningham2023sparseautoencodershighlyinterpretable}. \ac{saes} have recently been applied both in the mechanistic interpretability of \ac{llms} \citep{nanda2023progress, conmy2023towards, marks2024sparse}, and in deep learning more broadly \citep{math11081777}. \ac{saes} have been scaled to medium and large \ac{lms}, such as GPT4 \citep{templeton2024scaling, gao2024scalingevaluatingsparseautoencoders}.

\ac{saes} learn a reconstruction $\hat{\mathbf{x}}$ as a sparse linear combination of  features $\mathbf{y}_i \in \mathbb{R}^{d_{\text{s}}}$ for a given input activation $\mathbf{x} \in \mathbb{R}^{d_{\text{m}}}$ where $d_m$ is the \ac{llm}'s hidden size and:
\begin{equation}
d_{\text{s}} = k \, d_{\text{m}}, \quad \text{with } k \in \{2^n \mid n \in \mathbb{N}_+\}.
\end{equation}
The decoder element of the \ac{sae} is given as:
\begin{align} \label{eq:sae_recon} (\hat{\mathbf{x}} \circ \mathbf{f})(\mathbf{x}) = \mathbf{b}_d + \mathbf{W}_d\mathbf{f}(\mathbf{x})  \end{align}
where $\mathbf{b}_d \in \mathbb{R}^{d_{\text{m}}}$ is the bias term of the decoder, $\mathbf{W}_d$ is the decoder matrix with columns $\mathbf{v}_i \in \mathbb{R}^{d_{\text{m}}}$, and  $\mathbf{f}(\mathbf{x})$ denotes the feature activations, which are described by:
\begin{equation} \label{eq:sae_decomp} \mathbf{f}(\mathbf{x}) = \text{TopK}(\mathbf{W}_e(\mathbf{x} - \mathbf{b}_d) + \mathbf{b}_e) \end{equation}where $\mathbf{b}_e \in \mathbb{R}^{d_{\text{s}}}$ is the bias term of the encoder, $\mathbf{W}_e$ is the encoder matrix with columns $\mathbf{w}_i \in \mathbb{R}^{d_{\text{s}}}$, and the TopK activation function enforces sparsity following \citet{gao2024scalingevaluatingsparseautoencoders}.The loss function is the output's mean-squared error (MSE): \begin{equation} \label{eq:sae_loss} \mathcal L_{\text{}}= \norm{\mathbf{x} - \hat{\mathbf{x}}^2_2} \\ \end{equation}

\subsubsection{Embedders}
As a baseline, we replicate the embedding methodology of \citet{vamvourellis2023company}, and obtain embeddings for company descriptions. In particular, we use their three best performing embedders for our evaluations and downstream tasks:

\begin{enumerate}
    \item BERT: Pre-training of Deep Bidirectional Transformers for Language Understanding \citep{devlin2019bertpretrainingdeepbidirectional}.
    \item Sentence-BERT (SBERT): Building on BERT, SBERT improves latency substantially \citep{DBLP:journals/corr/abs-1908-10084} and encodes meaning on the more abstract sentence level.
    
    \item PaLM-gecko: Pathways Language Model (PaLM) \citep{chowdhery2022palmscalinglanguagemodeling}.

\end{enumerate}

\section{Methodology}

\subsection{Dataset}
Publicly listed companies in the US submit annual reports to the SEC, which include information on a company’s operations, such as product specifications, subsidiaries, competition, and other financial details\defcitealias{sec_form10k}{SEC, 2023}\citepalias{sec_form10k}. Due to the closed-source nature of GICS classifications, we use SIC-codes and the industry/major division categorization\footnote{The first 3 digits of the SIC code splits companies into 12 industry/major-divisions, referred to hereafter as BISC (Broader Industry Sector Code) \citep{SICmanual645}.} (BISC). Next, we tokenize company descriptions and preprocess them (Appendix~\ref{appendix:data}), resulting in a final dataset of 27,888 reports from 1996 to 2020.

\subsection{Feature summing}
In this work, we face the challenge of comparing sparse feature sequences of arbitrary lengths, where best practices are not well-established, though max-pooling has been proposed as a baseline for feature aggregation \citep{Anthropic}. However, motivated by the specific demands of financial sequence modeling, we propose an alternative, employing sparse feature summing across tokens. This method provides a magnitude-scaled count of the frequency with which a feature appears within a sequence, reflecting both the number of tokens on which a feature is active and its intensity \citep{lan2024sparse}.

Our approach is inspired by analogous methodologies in literature. For example, \citet{Loughran2009-LOUAWI-2} highlight the value of summing word counts in financial text analysis to derive domain insights. 

We sum sparse features, across tokens, from an \ac{sae} \citep{eleutherai_sae_llama} applied to layer 30 (occurring at 90\% of model depth). At this layer, we capture relevant features from preceding layers via the skip connection \citep{vaswani2017attention}, but not the logit-related features that tend to occur at the very last layers \citep{ghilardi-etal-2024-accelerating}.

The skip connection ensures that a single SAE captures the entire residual stream \citep{longon2024interpreting}, inherently including information from all preceding layers, thus ensuring that the summed sparse features represent a comprehensive aggregation of the model's internal representation of a company description.
\begin{figure}[t]
  \includegraphics[width=\columnwidth]{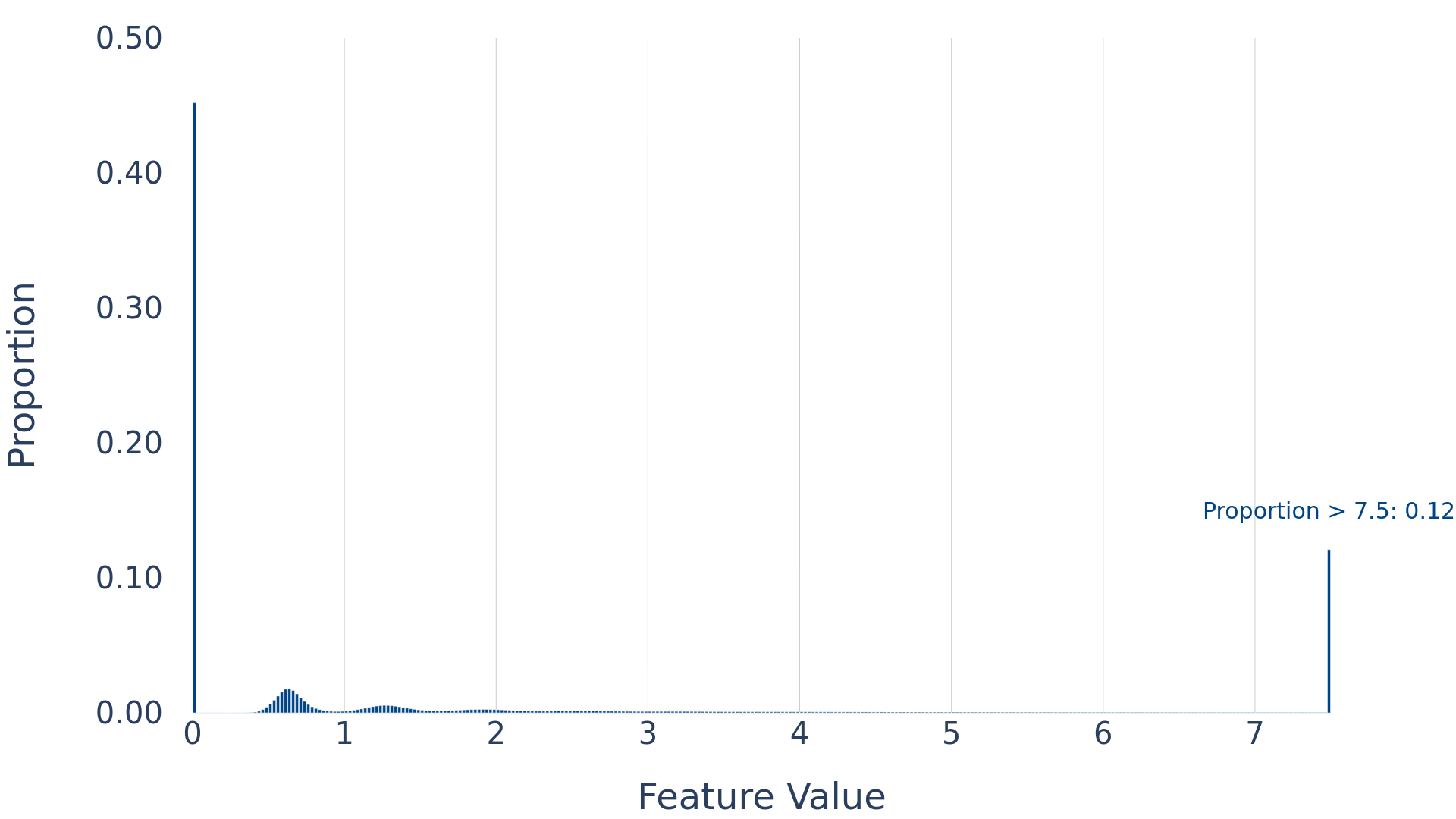}
  \caption{ Distribution of summed feature activations.}
  \label{fig:experiments}
\end{figure}
We analyze summed sparse features, and observe an interesting exponential decay pattern in feature activation frequencies (Figure~\ref{fig:experiments}).

Figure~\ref{fig:experiments} highlights the sparsity of \ac{llm} latent features -- even when these are summed across thousands of tokens -- motivating feature summing as an approach. In this context, a single active feature has, on average, before summing, a value of $\approx{0.7}$ (the first bulge).

This method also addresses a limitation in using embeddings \citep{vamvourellis2023company}, which require equal-length sequences for comparison. By focusing on cumulative feature occurrences, summed sparse features enable comparisons between sequences of arbitrary lengths, offering greater flexibility for analyzing variable-length financial datasets.

\subsection{Clustering}
\label{sec:clustering}
We benchmark our sparse features against embeddings and SIC/BISC-codes, where each SIC/BISC-code is its own cluster.

Each clustering method group \( G_k \) represents a distinct grouping methodology (i.e. \( G_{\mathrm{CD}} \) uses the cosine distance metric in our Sparse Features, while \( G_{\mathrm{BERT}} \) is based on the BERT embedders).

Within each model group \( G_k \), clusters are generated independently for each year from 1996 to 2020. Thus, \( G_k \) is formally structured as a set of yearly clustering outcomes:
\[
G_k = \left\{ G_k^{(y)} \mid y \in \{1996, 1997, \dots, 2020\} \right\},
\]

\noindent
where \( G_k^{(y)} \) is the set of clusters formed in year \( y \): 

\[ G_k^{(y)} = \{  C^{(y)}_1, C^{(y)}_2, \dots, C^{(y)}_n \},\]
where $C^{(y)}_i$ $\subseteq \{ \text{Companies in year } y \}$.  Each cluster $C^{(y)}_i$ contains a unique subset of companies active in year \( y \), ensuring that clusters are independent across different years.

To evaluate each clustering model, we compute the mean intra-cluster correlation $\text{MC}(G_k^{(y)})$:

\footnotesize
\[ \text{MC}(G_k^{(y)}) = \frac{1}{|G_k^{(y)}|} \sum_{C^{(y)}_i \in G_k^{(y)}} \frac{1}{|C^{(y)}_i|} \sum_{(a, b) \in C^{(y)}_i} \rho(a, b), \]
\normalsize
\noindent
where \( \rho(a, b) \) denotes the Pearson correlation of the logged monthly returns for companies \( a \) and \( b \) for the given year $y$. This metric quantifies the coherence of stock returns within clusters, providing a measure of how meaningful the cluster is.

We define the \textbf{overall mean correlation} (our main evaluation metric) of cluster groups \( G_k \) across years as:

\(\text{MC}(G_k) = \frac{1}{|\mathcal{Y}|} \sum_{y \in \mathcal{Y}} \text{MC}(G_k^{(y)})  \text{, where } \mathcal{Y} = \{1996, \dots, 2020\}.
\).

\subsubsection{Clustering sparse features}
Sparse features lack the locality and smoothness of embeddings \citep{KirosZS15, Bischke19} to define reliable similarity metrics. For instance, the TopK activation function \citep{gao2024scalingevaluatingsparseautoencoders} introduces sparsity, but with a strong discontinuity (truncates all features not in the top 128). 

To overcome these limitations, we apply Principal Component Analysis (PCA) to the raw features\footnote{We fit PCA globally across 1996–2020 for consistent eigenvectors, $n_{\text{components}}=4000$ captures 89.92\% variance.}. PCA mitigates the impact of non-activating features by reducing dimensionality, and retains only the most informative feature directions. Furthermore, PCA expedites our computations.

To cluster the PCA-transformed sparse features, we adopt the graph-theoretic framework of \citet{Bonanno_2004}, employing Minimum Spanning Trees (MSTs) to extract hierarchical structures from financial data. A fully connected graph is constructed with edge weights representing a particular distance metric. The MST encodes a subdominant ultrametric, with ultrametric distance defined by the maximum edge weight on the unique path between two nodes\footnote{To enforce the ultrametric property, we employ single-linkage hierarchical clustering, which groups nodes by iteratively merging the pair of clusters with the smallest maximum distance between any two points. This process satisfies the ultrametric inequality \((d_{ij} \leq \max(d_{ik}, d_{kj}))\) by construction.}. We remove edges above a specified weight level, defining this as the "cut-off threshold" ($\theta$), generating clusters directly from the MST. This eliminates the need for additional clustering steps, ensuring stable and interpretable results consistent with \citet{Bonanno_2004}.

\textbf{Cosine Distance}: We define the normalized cosine distance between our PCA-transformed sparse features as \(CD \), which we use for clustering. The resulting clusters are denoted as \( G_{\text{CD}} \). This metric measures dissimilarity, which captures angular separation rather than absolute magnitude differences \citep{Zafarani-Moattar2021}. For each pair of companies \( i \) and \( j \) such that both companies belong to the same year\footnote{Note that we define pairs (i,j), ensuring that company i and company j are only compared within the same year.}, the cosine similarity is computed as:
\[
S_{i,j} = \frac{\mathbf{g}_i \cdot \mathbf{g}_j}{\|\mathbf{g}_i\| \|\mathbf{g}_j\|}
\]
where \( \mathbf{g}_i \) and \( \mathbf{g}_j \) are the PCA-transformed feature vectors, \( \mathbf{g}_i \cdot \mathbf{g}_j \) denotes the dot product, and \( \|\mathbf{g}_i\| \) represents the Euclidean norm (\( L^2 \)-norm).

The cosine distance is then given by:
\[
d_{\cos}(i, j) = 1 - S_{i,j}
\]
We then normalize the cosine distance\footnote{Normalizing cosine-based distances can enhance the performance of clustering algorithms \citep{9655880}.}, defining the normalized distance function as $CD$. $CD$ is used to determine the edge weights of the Minimum Spanning Tree (MST), and we apply a cut-off threshold $\theta$ to prune high-weight edges. The resulting connected components define the clusters \( G_{\text{CD}} \)\footnote{We also refer to \( G_{\text{CD}} \) as \( G_{\text{Sparse\_Features}} \) in our paper.}.

\textbf{Cut-off $\theta$ calibration}:
To determine the MST cut-off threshold $\theta$ for $G_{\text{CD}}$, we initially apply a two-fold temporal cross-validation scheme: $\theta$ is chosen to maximize the average intra-cluster correlation across two time periods covering 25\% and 50\% of our dataset. We define this as $G_{\text{CD}}$ \footnote{See Appendix \ref{appendix:clustering sae} for the optimization of $G_\text{CD}$'s cutoff.}.

We ablate this choice by introducing a rolling variant. A separate $\theta_y^\star$ is chosen for each year $y$, based only on a five-year rolling lookback window:
\[
  \theta_y^\star
  = \arg\max_{\theta\in\{-4.5,-4.4,\dots,-1.0\}}
      \frac{1}{5}\sum_{s=y-5}^{y-1} \mathrm{MC}^{(s)}(\theta),
\]
We rebuild \(G_{\mathrm{CD}}^{(y)}\) with \(\theta_y^\star\), and report \(\mathrm{MC}^{(y)}(\theta_y^\star)\) as the yearly mean correlation statistic for each year \(y=2001,\dots,2020\); earlier years serve only as the look-back window. We define this rolling setup as $G_\text{CDR}$, for results see Appendix \ref{appendix:clustering rolling}, which confirms the robustness of our sparse‐feature clusters under strict out-of-sample evaluation.

\subsubsection{Clustering embeddings}
Following \citet{vamvourellis2023company}, each of the embedders discussed above is employed to define a unique clustering method group: (a) $G_\text{BERT}$; (b) $G_\text{SBERT}$; and (c) $G_\text{PaLM-gecko}$\footnote{We collectively refer to
$G_\text{BERT}$, $G_\text{SBERT}$, and $G_\text{PaLM-gecko}$ as $G_\text{Embedders}$ for simplicity and to streamline discussion.} (details in Appendix~\ref{appendix:emb}).

The SIC/BISC families are clusters by definition, and hence don't require further calibration.

\subsection{Pairs trading}
Our downstream task is pairs trading – a type of statistical arbitrage strategy that typically assumes a long-run equilibrium relationship between two stocks \citep{261eac3cf9934603a0fbb6bd2308462b}. We begin by splitting the dataset into an in-sample period (Jan 2002–Dec 2013) and an out-of-sample period (Jan 2014–Dec 2020), with clusters $G_k$ such that $k \in \{\text{Embedders, Sparse\_Features, SIC, BISC}\}$.

\normalsize
The pairs trading strategy consists of:

\begin{enumerate}
    \item \textbf{Pre-selection:} For each cluster \(C_i \in G_k\), stock pairs are filtered if the Pearson correlation of their monthly logged returns exceeds 0.95 during the in-sample period.
    \item \textbf{Co-integration Testing:} An Engle-Granger co-integration test is conducted on stock prices (Jan 2002–Dec 2013) of pre-selected pairs using the Augmented Dickey-Fuller (ADF) statistic to assess the stationarity of the residual spread. Pairs with a p-value below 0.01 are considered co-integrated.
    \item \textbf{Trading:} The identified co-integrated pairs for each $G_k$ are evaluated out-of-sample\footnote{See Appendix~\ref{appendix:trading details} for trading logic details} (Table~\ref{tab:clustering_performance}). We assess co-integration effectiveness within each method group $G_k$ via the entire portfolio's Sharpe ratio\footnote{The Sharpe Ratio quantifies risk-adjusted returns, measuring excess return per unit of risk \citep{Guasoni_2018, Peters_2011}.}.

\end{enumerate}

\subsection{Interpretability}
\normalsize\selectfont
We show interpretability over a sample of 1000 features across 300 clusters. Clusters are formed using cosine distance, which can be interpreted as parallelism between the feature vectors (feature proportionality). There is no linear mapping between features and cosine distance (Appendix \ref{sec:linear-distance-proof}), hence, we adopt an activation patching framework \citep{zhang2024activation} with respect to  cosine distance. This means that we obtain an interpretation of a cluster using the features that have the largest impact on cosine distance across the cluster when they are zeroed out (set to 0).

We define the importance of feature $i$ as the total absolute variation in cosine distance across the cluster when feature $i$ is zeroed out. Let $g_i, g_j$ be PCA-transformed feature vectors $i, j$. Moreover, let $g^{z}_i,g^{z}_j$ be the same vectors with feature $z$ set to 0 before applying the PCA. We define the absolute impact on the cosine distance of feature $z$:

\small
\[   
imp(z) = \sum_{i, j}^{cluster}\mid CD(g_i,g_j) - CD(g^{z}_i,g^{z}_j) \mid . 
\]
\normalsize

There are 2 necessary conditions for an interpretation of a cluster to be valid:

\begin{enumerate}
    \item \textbf{Sparsity} There are n = 131,072 features, and we need to interpret a cluster using only a small subset of $k << n$ \textit{important} features. 

    \item \textbf{Interpretability:} The sparse features that we use need to be interpretable on our dataset.
\end{enumerate}
\normalsize\selectfont
To obtain the set of \emph{important} sparse features that constitutes the interpretation of a cluster, let \(F\) be the full set of \(n\) characteristics and define \emph{impact} of a subset of features \(S \subseteq F\) as follows:

\[
\mathrm{IMP}(S) = \sum_{z \in S} \mathrm{imp}(z).
\]
Then the set of \emph{important} features, \(S^*\), is given by

\small
\[
S^* = \underset{S \subseteq F}{\arg \min} \bigl|S\bigr|
\quad 
\text{subject to}
\quad
\mathrm{IMP}(S) \geq \mathrm{IMP}\bigl(F \setminus S\bigr).
\]
\normalsize
\(S^*\) is the smallest subset of features whose total impact on cosine distance in the cluster equals or exceeds that of the remaining features. We populate \(S^*\) by adding the most important feature in $F \setminus S$ to $S$ until $I\!M\!P(S) \geq I\!M\!P(F \setminus S)$.

We interpret our \emph{important} features using an auto-interpretability pipeline. First, the Gemini 2 Flash language model is prompted to explain a feature given examples of when the feature activates and when it does not. Then, the model predicts latent activations for new sentences based on its prior explanations (\textit{fuzzing}). Interpretability is measured as the success rate in fuzzing.

While there is no benchmark for the interpretability of Llama 3.1 8B sparse features, we compare with the closest benchmark in the literature: Gemma 2 9B on the "Red Pajama" and "The Pile" datasets \citep{paulo2024auto-interp}.

\section{Results}
\begin{figure*}[t]
    \centering
    \includegraphics[width=1\linewidth]{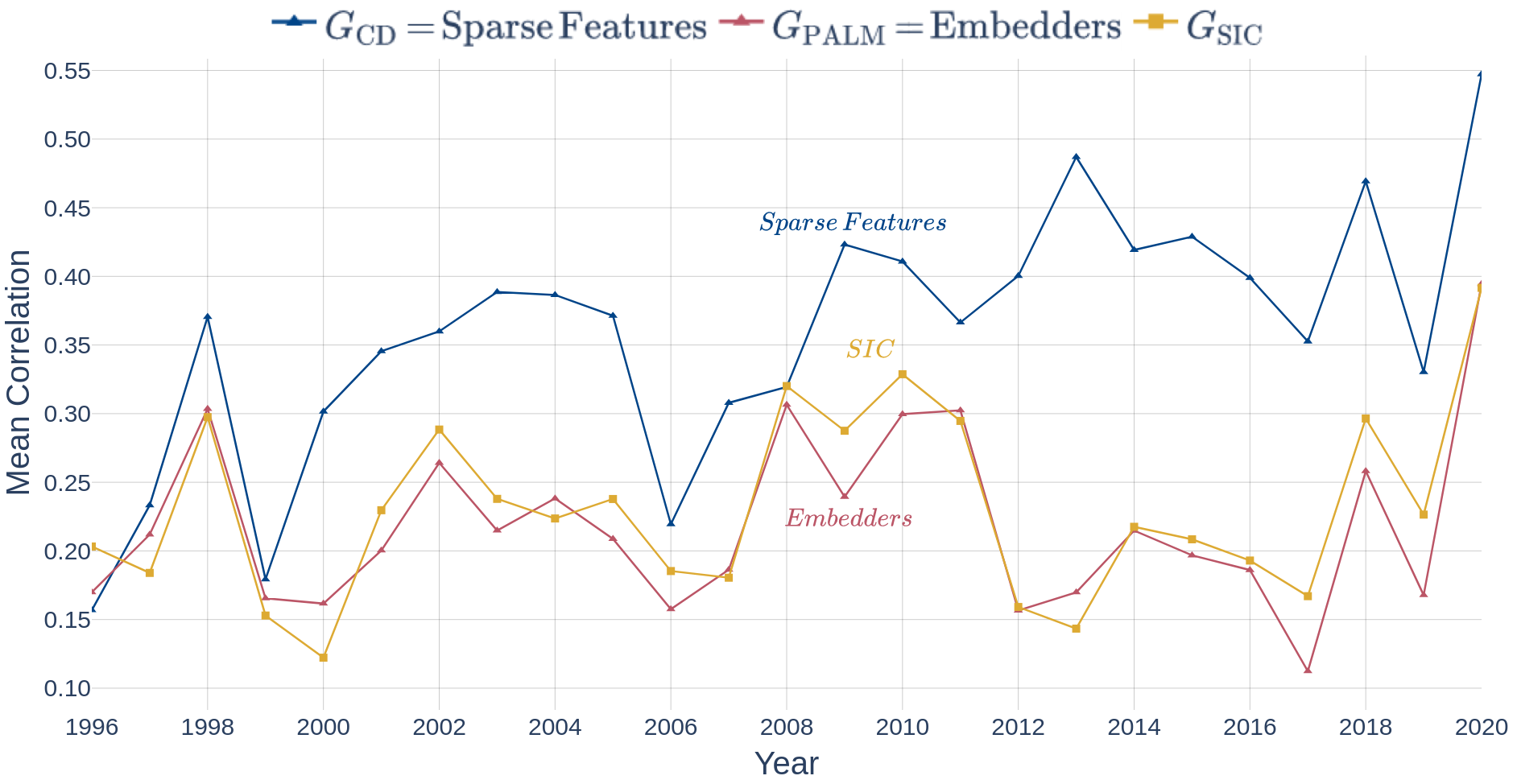}
    \caption{Overall Mean Correlation ($\text{MC}(G_k)$) of $G_\text{CD}$ (Normalized Cosine Distance Cluster Group) vs PaLM vs SIC Benchmarks between 1996-2020.
    Note that we use PaLM and SIC-codes for comparison, as they have the highest $\text{MC}(G_k)$ among the embedding-based and traditional benchmark groups, respectively.}
    \label{fig:Clustering-Result}
\end{figure*}

\subsection{Clustering results}

For each clustering method group $G_k$, we evaluate their $\text{MC}(G_k)$, and Sharpe Ratios (see Table~\ref{tab:clustering_performance} and Figure~\ref{fig:Clustering-Result}). The results demonstrate that clusters derived from our Sparse Features significantly outperform Embeddings, SIC-codes and BISC in terms of clustering similar companies.
\begin{table}[h]
\centering
\scriptsize
\setlength{\tabcolsep}{6pt} 
\small 
\renewcommand{\arraystretch}{1.2} 
\begin{tabular}{@{}lcc@{}}
\toprule
\textbf{Clustering Group $\mathbf{(G_k)}$} & \textbf{$\textbf{MC}\mathbf{(G_k)}$} & \textbf{Sharpe Ratio} \\ 
\midrule
\multicolumn{3}{c}{Our Contribution} \\
$\mathbf{G_\text{CD}}$ & $\mathbf{0.359}$ & $\mathbf{12.18}$\\
 $\mathbf{G_\text{CDR}}$& $\mathbf{0.385}$&$\mathbf{9.69}$\\ 
\midrule
\multicolumn{3}{c}{Embedding Benchmark} \\ 
\midrule
$G_\text{BERT}$ & 0.198 & 7.58\\
$G_\text{SBERT}$ & 0.219 & 7.69\\
$G_\text{PaLM-gecko}$ & 0.219 & 10.57\\
\midrule
\multicolumn{3}{c}{Traditional Benchmark Cluster Groups} \\ 
\midrule
$G_\text{SIC}$ & 0.231 & 9.70\\ 
$G_\text{BISC}$ & 0.187 & 7.58\\ 
Population\footnotemark& 0.161 & $\mathbf{-}$\\ 
\bottomrule
\end{tabular}
\caption{ Performance comparison between different clustering groups (averaged across 1996-2020).}
\label{tab:clustering_performance}
\end{table}
\footnotetext{Population group represents $\text{MC}{(G_k)}$ on the full dataset.}

\FloatBarrier

\subsection{Pairs trading results }
Sharpe ratios (risk-adjusted profits) were recorded for evaluation in backtesting. Within pairs trading, \citet{SungjuHong} find pairs with higher fundamental similarity outperform those with weaker economic ties by reducing non-convergence risk. In line with these findings, our clustering approach can outperform Embedders and Traditional Classifications in Sharpe Ratio (Table~\ref{tab:clustering_performance}), suggesting it may capture more fundamental company similarities.

\FloatBarrier
\subsection{Interpretability results}
\begin{table}[h]
    \centering
    \setlength{\tabcolsep}{6pt} 
\small 
\renewcommand{\arraystretch}{1.2}
    \begin{tabular}{@{}lc@{}}
        \toprule
        \multicolumn{2}{c}{\textbf{Interpretability}}\\
        \midrule
        \multicolumn{2}{c}{Our Contribution} \\
        Top 1\% Features ($G_\text{CD}$) \footnotemark& \textbf{80\%} \\
        Top 1\% Features ($G_\text{CDR}$) & \textbf{77\%} \\
        Average Feature & 62\% \\
        \midrule
        \multicolumn{2}{c}{Interpretability Benchmarks (Gemma 2 9B)} \\
        The Pile & 76\% \\
        Red Pajama & 76\% \\
        \midrule
        \multicolumn{2}{c}{Random Interpretation Baseline} \\
        Fuzzing Score & 51\% \\
        \bottomrule
    \end{tabular}
    \caption{Interpretability of SAE Features.}
    \label{tab:sae_interpretability}
\end{table}
With regards to our first interpretability requirement, sparsity, we measure what percentage of features are \emph{important} per cluster (Appendix~\ref{appendix:sparsity}), and find that the median cluster is very sparse with only $5\%$ of \emph{important} features. 
\footnotetext{
Top 1\% features are important for more clusters than the remaining 99\%, they are not the top 1\% for interpretability.}

\begin{figure}[H]
    \centering
    \includegraphics[width=1\linewidth]{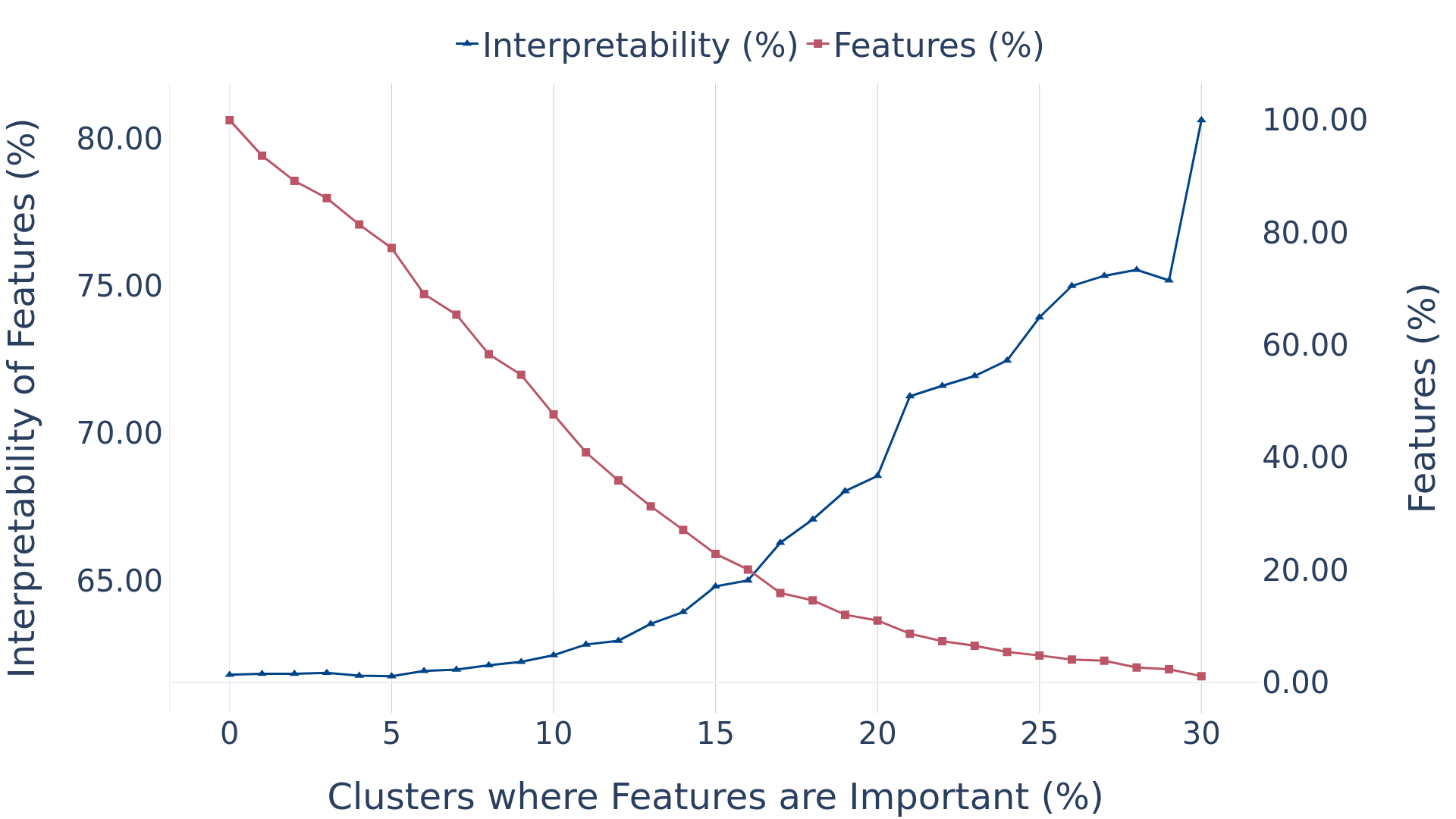}
    \caption{{Interpretability Score of Features by Percentage of Clusters ($G_\text{CD}$) where Features are Important. Data selected between 100\% (all features) and 1\%.}}
    \label{fig:Interpretability by importance}
\end{figure}

In terms of interpretability, we observe that most features are interpretable (Table~\ref{tab:sae_interpretability}). Moreover, features that are \emph{important} across multiple clusters, those we most want to interpret, also tend to be more interpretable (Figure~\ref{fig:Interpretability by importance}). In particular, \textit{top $1\%$ features} (features in the first percentile for the amount of clusters they are important for) are $80\%$ interpretable.

\FloatBarrier

\begin{figure}[H]
    \centering
    \includegraphics[width=1\linewidth]{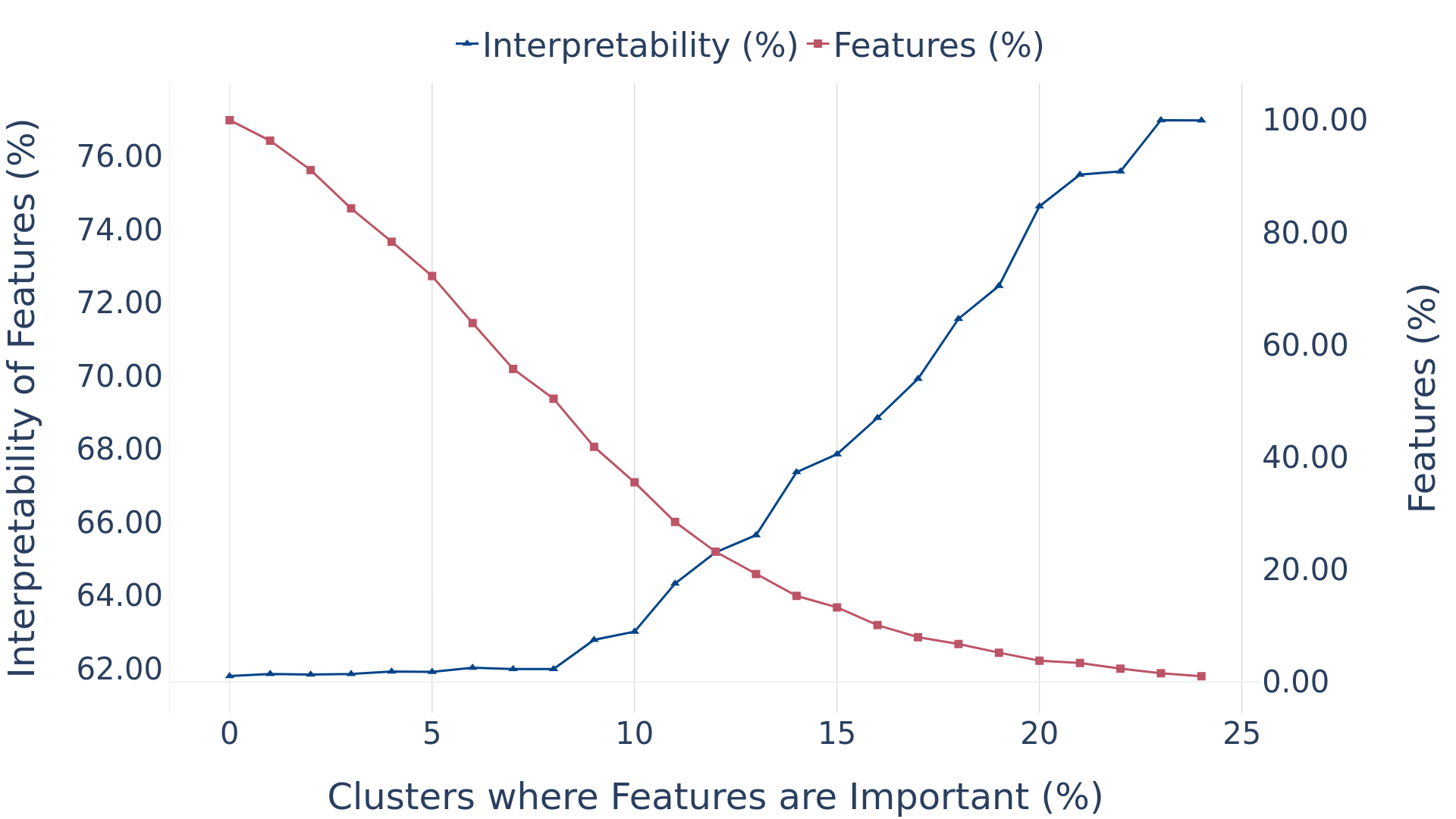}
    \caption{{Interpretability Score of Features by Percentage of Clusters ($G_\text{CDR}$) where Features are Important.}}
    \label{fig:Interpretability by importance (rolling)}
\end{figure}

Finally, we run the same experiments on the clusters constructed using the rolling cutoff (i.e. $G_\text{CDR}$), and our experiments yield similar results: \textit{top $1\%$ features} are $77\%$ interpretable. In terms of sparsity, the median cluster is very sparse with only $1\%$ of \emph{important} features (Appendix~\ref{appendix:sparsity}). The trend where more \textit{important} features are more interpretable also holds (see Figure~\ref{fig:Interpretability by importance (rolling)}).

\subsection{Limitations}
We do not fine tune embedders, \ac{saes}, or \ac{llms}. These could be exciting directions for future work. Reported Sharpe ratios should be interpreted cautiously as they may be sensitive to the choice of $\theta$, slippage, regime shifts, and finite-sample bias \citep{Lo2002StatisticsSharpe,bailey2012sharpe}.


%


\section{Conclusions}

We find that using \ac{sae} features is an effective and interpretable method for computing company similarity. Future work might explore applications in portfolio diversification and hedging strategies; optimizing trading strategies through fine-tuning $\theta$ and modeling shifts in economic regimes; extending the framework to other domains such as healthcare; or ablation studies such as replacing MST clustering with K-means.

\section{Acknowledgments}
We acknowledge and thank Nscale for providing the compute resources (8 AMD Mi250x GPUs) used for all SAE inference and most evaluations in this paper. We are especially grateful to Karl Havard for leading this partnership, Konstantinos Mouzakitis for his technical assistance, Brian Dervan for structuring our collaboration, and the entire Nscale team for their support.

We are grateful to  Vittorio Carlei at Qi4M for his knowledge and advice.

\bibliography{custom}

\appendix

\section{Data Preprocessing}
\label{appendix:data}
We consider 220,275 annual SEC reports from 1993 to 2020, ignoring any de-lists, accompanied by related meta-data on Company Name, Year, SIC-code, and CIK number (a unique SEC corporation identifier) \citep{sec_cik_lookup}. CIK numbers are mapped to their corresponding publicly traded ticker symbol, from which the monthly logged returns are retrieved via \citet{yahoofinance}.
We remove entries with missing or very short: company descriptions, ticker information, or monthly returns.
This leaves us with 27,888 reports.
We tokenize using Meta’s Llama 3 8B Tokenizer \citep{dubey2024llama3herdmodels}. We only retain companies that are consistently available for at least five years. In our analysis, we ignore pre-1996 data as the sample size is too small.
To refine the dataset further, we retain only annual reports with token counts within the context window.

\section{Clustering Embeddings}
\label{appendix:emb}
For BERT, we used \verb|bert-base-uncased| from the transformers library. For SBERT, we used \verb|all-MiniLM-L6-v2| from the \verb|sentence_transformers| library. For PaLM-gecko, we used \verb|textembedding-gecko@003| from the \verb|vertexai| library.

\textbf{Chunking:}
In our methodology, for both BERT and SBERT, we followed \citet{vamvourellis2023company} and implemented a chunking mechanism to accommodate the models' maximum token limit of 512. Specifically, company descriptions exceeding this limit were split into overlapping chunks of 512 tokens. The \verb|[CLS]| embeddings of these chunks were averaged to generate a single document embedding of 1536 tokens. For PaLM-Gecko, we leveraged its extended context window of 3072 tokens and directly processed the descriptions without chunking.

The pipeline below is optimised through Optuna’s Tree-structured Parzen Estimator (TPE) sampler for Bayesian hyperparameter optimization. The objective function maximizes $\text{MC}(G_k)$. This search is constrained to 150 trials and a maximum timeout of 9 hours to balance thoroughness and resource usage:

\textbf{Dimensionality Reduction with UMAP:}
Given the high dimensionality of the input embeddings (768-dimensional vectors derived from a BERT model), we first employ Uniform Manifold Approximation and Projection (UMAP) \cite{mcinnes2020umapuniformmanifoldapproximation} to reduce these high-dimensional textual embeddings to a lower-dimensional space, preserving both local and global data structures. We optimize three UMAP parameters to improve the quality of the downstream clustering: (a) \verb|n_components| (target dimensionality); (b) \verb|n_neighbors|; and (c) \verb|min_dist|. All embeddings are standardized and casted to \verb|float32| to ensure computational efficiency.

\textbf{Clustering with Spectral Clustering:}
After reducing dimensionality, we perform clustering using Spectral Clustering, which is capable of handling noise and complex cluster shapes, following \citet{vamvourellis2023company}. We first construct an affinity matrix from a k-nearest neighbors (KNN) graph of the UMAP outputs. Spectral Clustering then operates on this graph’s eigenstructure to form clusters. The number of clusters (\verb|n_clusters|) is tuned via Optuna, while the neighborhood size ($k$) is set to a constant of 5, following \citet{vamvourellis2023company}. 

\textbf{Temporal Cross-Validation:}
To evaluate the stability and temporal generalization of the resulting clusters, we employ temporal cross-validation. The dataset is split into chronological folds. This setup reduces temporal bias and assesses whether the identified cluster structure remains consistent over time. We used parallel processing to evaluate each fold.

\begin{table}[h]
\centering
\scriptsize
\setlength{\tabcolsep}{4pt} 
\renewcommand{\arraystretch}{1.2} 
\resizebox{\linewidth}{!}{ 
\begin{tabular}{@{}lccc@{}}
\toprule
\textbf{Embedder Cluster Group}  $(G_\text{embedder})$& \textbf{UMAP $n_\text{components}$} & \textbf{UMAP $n_\text{neighbors}$} & \textbf{UMAP} $\scriptsize{\text{min\_dist}}$\\ 
\midrule
$G_\text{BERT}$          & 7& 119&  0.109\\ 
$G_\text{SBERT}$         & 7& 79&  0.012\\ 
$G_\text{PaLM-gecko}$    & 6& 40&  0.120\\
\bottomrule
\end{tabular}
}
\caption{ Optimized UMAP Thresholds for Embedders}
\label{tab:Optimized UMAP Thresholds for Embedders}
\end{table}
\FloatBarrier
\begin{table}[h]
\centering
\scriptsize
\setlength{\tabcolsep}{4pt} 
\renewcommand{\arraystretch}{1.2} 
\resizebox{\linewidth}{!}{ 
\begin{tabular}{@{}lll}
\toprule
\textbf{Embedder Cluster Group}  $(G_\text{embedder})$& \textbf{Spectral $n_\text{clusters}$}&\textbf{Spectral $n_\text{neighbors}$}\\ 
\midrule
$G_\text{BERT}$          & 10&5\\ 
$G_\text{SBERT}$         & 49&5\\ 
$G_\text{PaLM-gecko}$    & 27&5\\
\bottomrule
\end{tabular}
}
\caption{\small Optimized Spectral Clustering Thresholds for Embedders}
\label{tab:Optimized Spectral Clustering Thresholds for Embedders}
\end{table}
\FloatBarrier
\section{Clustering Sparse Features}
\label{appendix:clustering sae}
\FloatBarrier
\begin{figure}[H]
    \centering
    \includegraphics[width=1\linewidth]{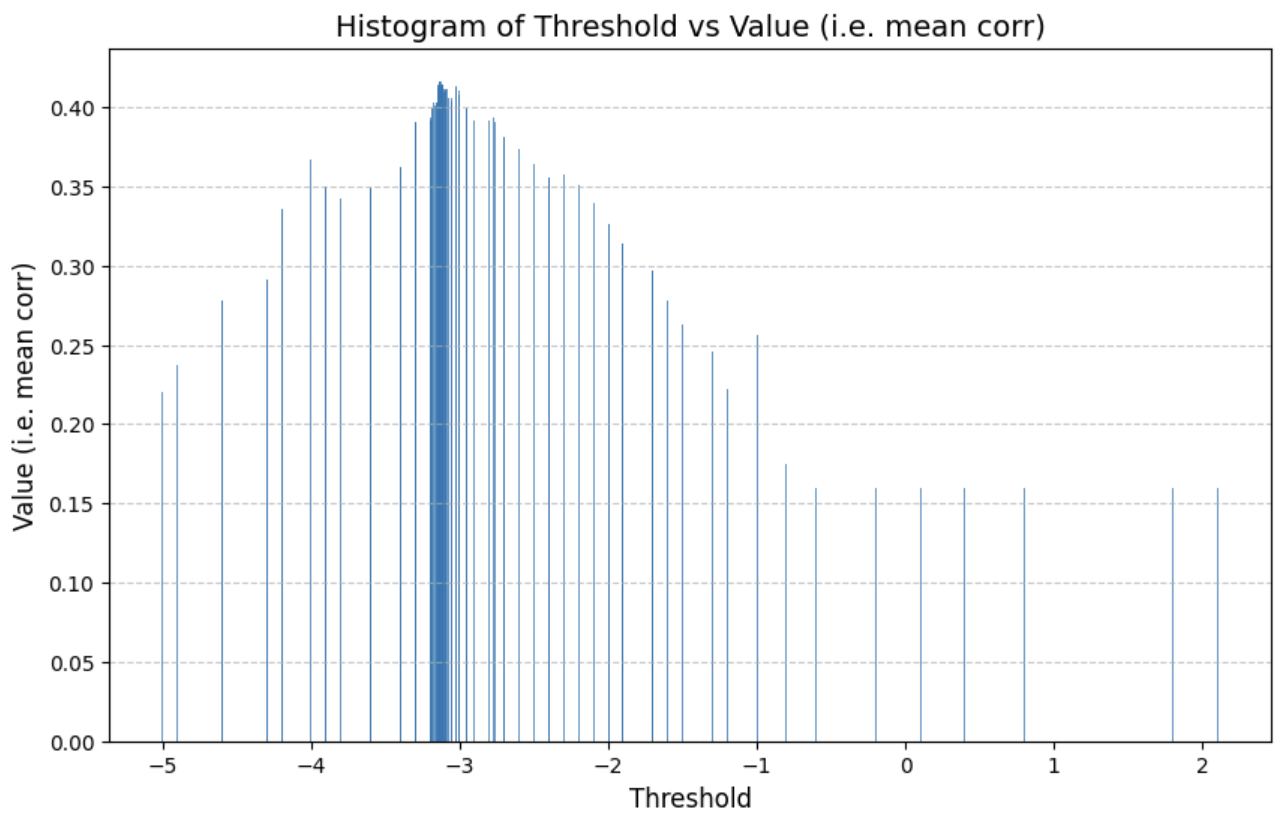}
    \caption{Optuna Study – Histogram of Sparse Features' MST cutoff thresholds. Maximizing Threshold = -3.130.}
    \label{fig:Optuna Study for Sparse Features}
\end{figure}
\FloatBarrier
Figure~\ref{fig:Optuna Study for Sparse Features} plots the distribution of candidate MST cut-off values \(\theta\) (x-axis) against their corresponding mean intra-cluster correlations (y-axis). The long right tail approaches the overall population mean correlation ($\approx{0.161}$) as \(\theta\) loosens, while bulk of high MeanCorr values sits to the left (lower \(\theta\)), reflecting tighter distance threshold groups similar firms.

\section{Clustering Sparse Features OOS with Rolling Frame}
\label{appendix:clustering rolling}
\begin{figure}
    \centering
    \includegraphics[width=1\linewidth]{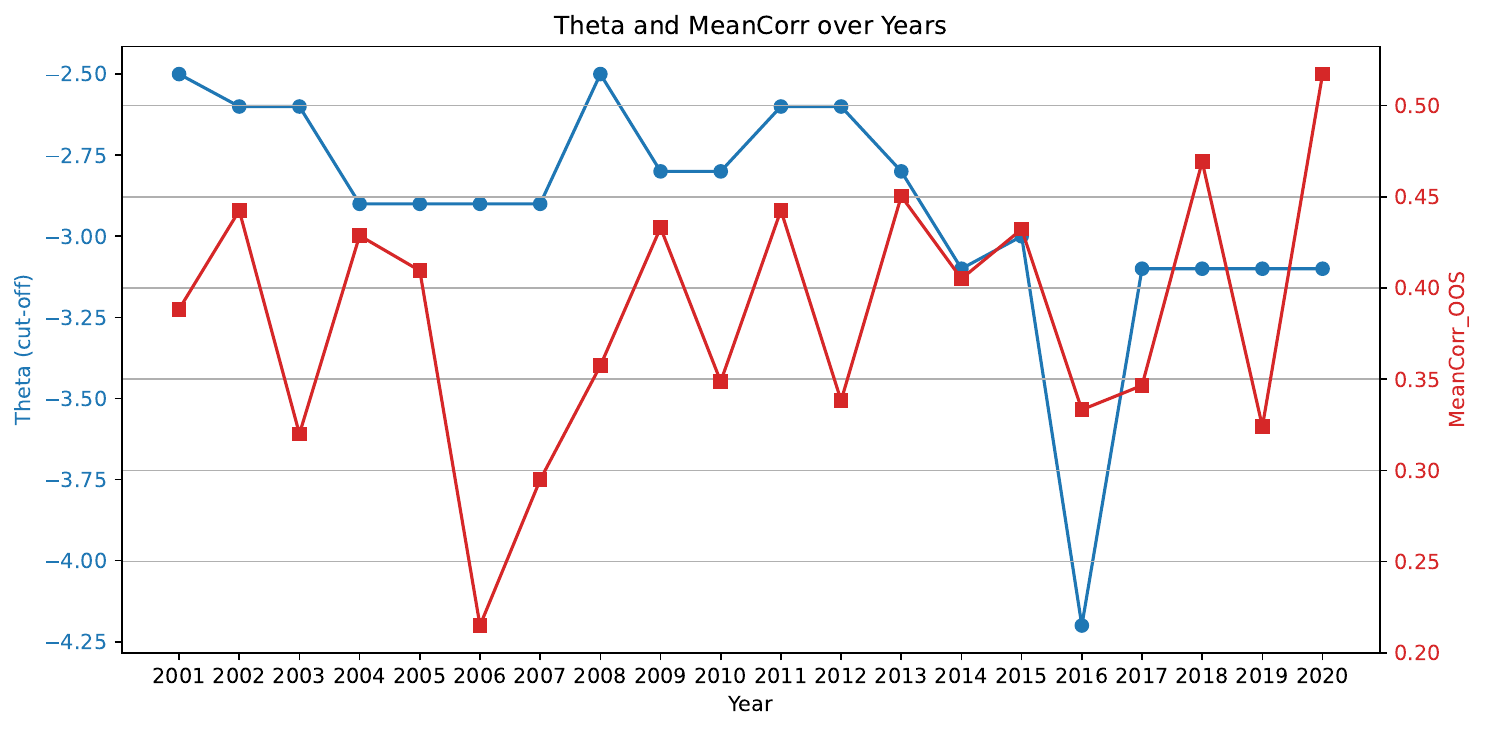}
    \caption{
Walk-forward tuning results for the sparse-feature ($G_\text{CDR}$). 
    \textcolor{blue}{Blue (left axis)}: optimal MST edge-weight cut-off \(\theta_y^\star\) obtained from the preceding five-year rolling window. 
    \textcolor{red}{Red (right axis)}: resulting out-of-sample per-year mean intra-cluster correlation \(\text{MC}^{\text{OOS}}_y\). }
    \label{fig:rollingTheta}
\end{figure}
In terms of results, the forward rolling variant achieves a higher overall mean correlation of $\text{MC}(G_{\text{CDR}}) = 0.391$, compared to the temporal fold result of $\text{MC}(G_{\text{CD}}) = 0.359$. As shown in Figure~\ref{fig:rollingTheta}, the optimal cut-off $\theta_y^\star$ evolves smoothly over time, while the out-of-sample mean intra-cluster correlation remains between 0.30 and 0.46 in most years—peaking in 2020 when market-wide correlations surged during the COVID-19 crisis. These findings confirm the robustness of our sparse-feature clusters under forward-looking evaluation.

\section{Trading Details}
\label{appendix:trading details}
For each clustering‐based strategy \(G_k\), we simulate pair trades over the out‐of‐sample period 2014–2020 and record, for each business day \(t\), the total portfolio value \(V_{k,t}\). This series acts as the portfolio trajectory and is constructed as follows:
(1) On each business day \(t\), add realized PnL from any closed trades to cash. (2) Mark open positions to market and compute unrealized P\&L. (3) Set \( V_{k,t} = \text{cash}_t + \text{unrealized\_PnL}_t\) and append it to the portfolio trajectory series, which was subsequently used for Sharpe ratio calculations.

Following \citet{Miao_2014}, we assumed zero transaction costs, opening positions when the residual spread deviated beyond \(\pm 1\sigma\) its mean, and closing when the spread reverted to the mean. A stop-loss mechanism is triggered if the spread exceeds \(\pm 2\sigma\). We obtained stock price data via finance.

\section{Feature Sparsity Analysis}
\label{appendix:sparsity}
\begin{figure}[H]
    \centering
    \includegraphics[width=1\linewidth]{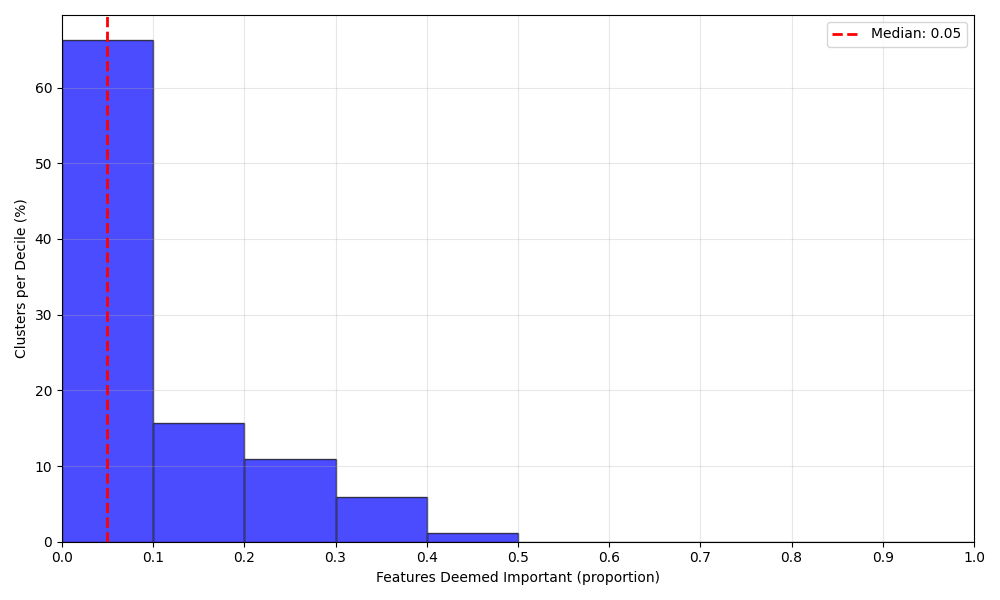}
    \caption{Distribution of the proportion of important features over clusters ($G_\text{CD}$).}
    \label{fig:Important faetures per cluster}
\end{figure}

\section{Feature Sparsity Analysis}
\label{appendix:sparsity}
\begin{figure}[H]
    \centering
    \includegraphics[width=1\linewidth]{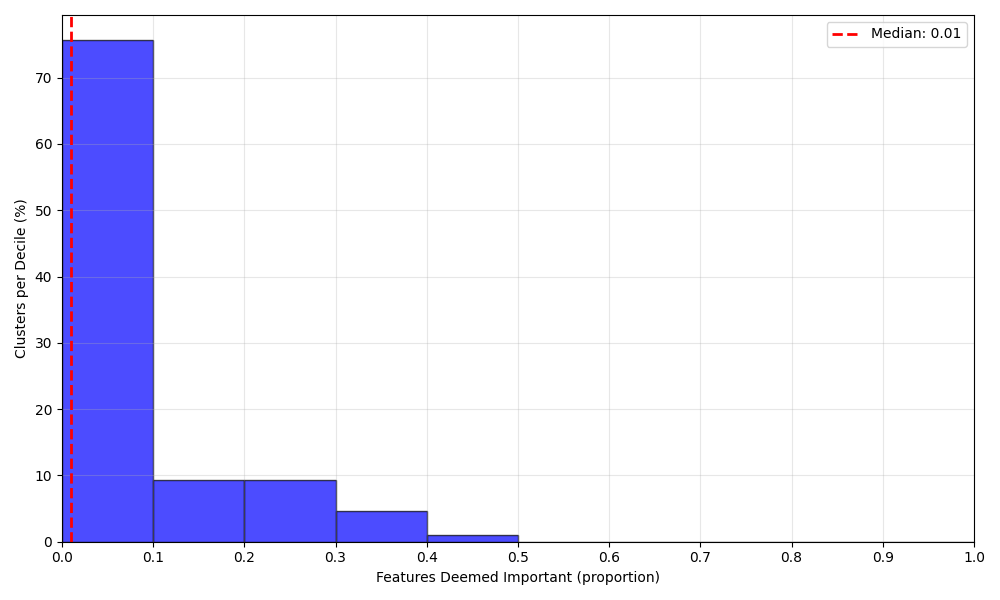}
    \caption{Distribution of the proportion of important features over clusters ($G_\text{CDR}$).}
    \label{fig:Important faetures per cluster (rolling)}
\end{figure}

\section{Why a Linear Distance Must Be Trivial}
\label{sec:linear-distance-proof}

\textbf{Claim.} If a function $d(\cdot,\cdot)$ on a vector space is both a 
\emph{distance function} (metric) and \emph{linear} in its arguments 
(plus symmetry), then $d(x,y)=0$ for all $x,y$. 

\begin{proof}
By the metric property, $d(z,z)=0$ for any $z$. Pick arbitrary vectors $x$ and $y$, and let $z = x+y$. Then
\[
0 
= d(z,z) 
= d(x+y,\;x+y).
\]
Assume $d$ is linear in the first argument and symmetric. By linearity on the first argument,
\[
d(x+y,\; x+y) \;=\; d(x,\;x+y)\;+\;d(y,\;x+y).
\]
By symmetry, $d(x,\;x+y) = d(x+y,\;x)$.  Applying linearity in the first argument again,
\[
d(x+y,\;x)
= d(x,\;x) + d(y,\;x)
= 0 + d(y,\;x),
\]
because $d(x,x) = 0$ from the metric property. Symmetry again gives $d(y,x) = d(x,y)$. Hence
\[
d(x,\;x+y) = d(x,y).
\]
Similarly, $d(y,\;x+y) = d(x,y)$. Therefore,
\[
d(x+y,\;x+y)
= 2\,d(x,y).
\]
But $d(x+y,\;x+y) = 0$, so $2\,d(x,y) = 0 \implies d(x,y)=0$ for all $x,y$.  
Thus, if a distance were to be linear, it would be zero for all elements x,y,
contradicting the usual requirement $d(x,y)=0 \iff x=y$ unless 
the entire space is collapsed.
\end{proof}

\end{document}